\documentclass[10pt,twocolumn,letterpaper]{article}

\usepackage{cvpr}

\cvprfinalcopy

\usepackage{times}
\usepackage{epsfig}
\usepackage{graphicx}
\usepackage{amsmath}
\usepackage{amssymb}
\usepackage{rotating}
\usepackage{lettrine}
\usepackage{color}
\usepackage{flushend}
\usepackage{framed}
\usepackage{authblk}

\def \ie {\emph{i.e.}\xspace}
\def \eg {\emph{e.g.}\xspace}
\def \etal {\emph{et al.}\xspace}

\def \X {{\mathrm X}}
\def \Xset {{\mathcal X}}
\def \N {{\mathbf N}}
\def \G {{\mathbf G}}
\def \Y {{\mathbf Y}}
\def \betastar {{\beta^\star}}

\def \Bset {{\mathcal B}}

\def \mysp {\hspace{4pt}}

\newcommand{\OURMETHOD}{L\&C\xspace}

\definecolor{antiquefuchsia}{rgb}{0.57, 0.36, 0.51}

\usepackage{enumitem}
\setlist[itemize]{%
labelsep=5pt,%
labelindent=0.4\parindent,%
itemindent=0pt,%
leftmargin=*,%
itemsep=-1pt, 
}

\usepackage[pagebackref=false,breaklinks=true,letterpaper=true,colorlinks,bookmarks=false]{hyperref}

\def\cvprPaperID{3659} %

\ifcvprfinal\fi
\begin{document}

\title{Link and code: Fast indexing with graphs and compact regression codes
} 

\author[$\dagger$]{Matthijs Douze}
\author[$\dagger$,$\star$]{Alexandre Sablayrolles}
\author[$\dagger$]{Herv\'e J\'egou}

\affil[ ]{\vspace{-3pt}$^\dagger$Facebook AI Research\; \quad \quad \quad $^\star$Inria\thanks{University Grenoble Alpes, Inria, CNRS, Grenoble INP, LJK.}}

\maketitle
\begin{abstract}
Similarity search approaches based on graph walks have recently attained outstanding speed-accuracy trade-offs, taking aside the memory requirements. In this paper, we revisit these approaches by considering, additionally, the memory constraint required to index billions of images on a single server. This leads us to propose a method based both on graph traversal and compact representations. We encode the indexed vectors using quantization and exploit the graph structure to refine the similarity estimation. 

In essence, our method takes the best of these two worlds: the search strategy is based on nested graphs, thereby providing high precision with a relatively small set of comparisons. At the same time it offers a significant memory compression. As a result, our approach outperforms the state of the art on operating points considering 64--128 bytes per vector, as demonstrated by our results on two billion-scale public benchmarks. 
\end{abstract}

\section{Introduction}
\label{sec:introduction}

Similarity search is a key problem in computer vision. It is a core component of large-scale image search~\cite{LCL04,PCISZ07}, pooling~\cite{SZ03} and semi-supervised low-shot classification~\cite{douze2017low}. Another example is classification with a large number of classes~\cite{he2015learning}. 
In the last few years, most of the recent papers have focused on compact codes, either binary~\cite{C02,GL11} or based on various quantization methods~\cite{JDS11,CGW10,BL14,ZQTW15}. Employing a compact representation of vectors is important when using local descriptors such as SIFT~\cite{L04}, since thousands of such vectors are extracted per image. In this case the set of descriptors typically requires more memory than the compressed image itself. Having a compressed indexed representation employing $8-32$ bytes per descriptor was a requirement driven by scalability and practical considerations. 

However, the recent advances in visual description have mostly considered description schemes~\cite{PLSP10,JDSP10,GWGL14} for which each image is represented by an unique vector, typically extracted from the activation layers of a convolutional neural network~\cite{BL15,TSJ16}. The state of the art in image retrieval learns the representation end-to-end~\cite{GARL16,RTC16} such that cosine similarity or Euclidean distance reflects the semantic similarity. The resulting image descriptors consist of no more than a few hundred components. 

In this context, it is worth investigating approaches for nearest neighbor search trading memory for a better accuracy and/or efficiency. An image representation of 128 bytes is acceptable in many situations, as it is comparable if not smaller than the meta-data associated with it and stored in a database. 
While some authors argue that the performance saturates beyond 16 bytes~\cite{babenko2017annarbor}, the best results achieved with 16 bytes on the Deep10M and Deep1B datasets do not exceed $50\%$ recall at rank 1~\cite{babenko2016efficient,douze2016polysemous}. While going back to the original vectors may improve the recall, it would require to access a slower storage device, which would be detrimental to the overall efficiency. 

\begin{figure}[t]
\centering
\includegraphics[width=0.55\linewidth]{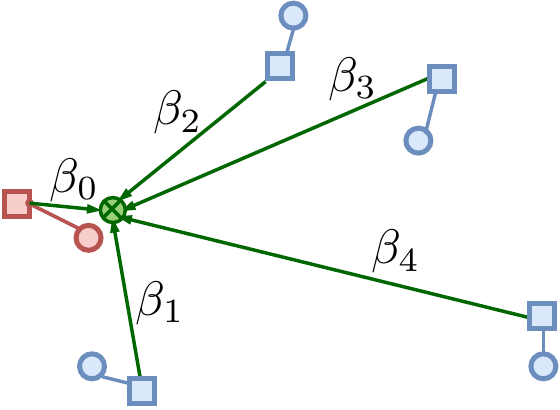}
\caption{Illustration of our approach: we adopt a graph traversal strategy~\cite{malkov2016efficient} that maintains a connectivity between all database points. Our own algorithm is based on compressed descriptors to save memory: each database vector (circle) is approximated (square) with quantization. We further improve the estimate by regressing each database vector from its encoded neighbors, which provides an excellent representation basis. The regression coefficients $\boldsymbol{\beta}=[\beta_0,\dots,\beta_k]$ are selected from a codebook learned to minimize the vector reconstruction error. 
\label{fig:toy_illustration}}
\vspace{-10pt}
\end{figure}

One the opposite, some methods like those implemented in FLANN~\cite{ML14} consider much smaller datasets and target high-accuracy and throughput. The state-of-the-art methods implemented in NMSLIB~\cite{BN13} focuses solely on the compromise between speed and accuracy. They do not include any memory constraint in their evaluation, and compare methods only on small datasets comprising a few millions vectors at most. Noticeably, the successful approach by Malkov \etal~\cite{malkov2014approximate,malkov2016efficient} requires both the original vectors and a full graph structure linking vectors. This memory requirement severely limits the scalability of this class of approaches, which to the best of our knowledge have never been scaled up to a billion vector. 

These two points of views, namely compressed-domain search and graph-based exploration, consider extreme sides of the spectrum of operating points with respect to memory requirements. While memory compactness has an obvious practical advantage regarding scalability, we show in Section~\ref{sec:analysis} that HNSW (Hierarchical Navigable Small Worlds~\cite{malkov2014approximate,malkov2016efficient}) is significantly better than the Inverted Multi-Index (IMI)~\cite{BL15pami} in terms of the compromise between accuracy and the number of elementary vector comparisons, thanks to the effective graph walk that rapidly converges to the nearest neighbors of a given query. 

We aim at conciliating these two trends in similarity search by proposing a solution that scales to a billion vectors, thanks to a limited memory footprint, and that offers a good accuracy/speed trade-off offered by a graph traversal strategy. For this purpose, we represent each indexed vector by i) a compact representation based on the optimized product quantization (OPQ)~\cite{GHKS13}, and ii) we refine it by a novel quantized regression from neighbors. This refinement exploits the graph connectivity and only requires a few bytes by vector. Our method learns a regression codebook by alternate optimization to minimize the reconstruction error. %

The contributions of our paper consist of a preliminary analysis evaluating different hypotheses, and of an indexing method employing a graph structure and compact codes. Specifically, 
\begin{itemize}
\item We show that using a coarse centroid provides a better approximation of a descriptor than its nearest neighbor in a typical setting, suggesting that the first approximation of a vector should be a centroid rather than another point of the dataset~\cite{babenko2017annarbor}.  
We also show that a vector is better approximated by a linear combination of a small set of its neighbors, with fixed mixing  weights obtained by a close-form equation. This estimator is further improved if we can store the weights on a per-vector basis.
\item %
We show that HNSW offers a much better selectivity than a competitive method based on inverted lists. This favors this method for large representations, as opposed to the case of very short codes (8--16 bytes). 
\item We introduce a graph-based similarity search method with compact codes and quantized regression from neighbors. It achieves state-of-the-art performance on billion-sized benchmarks for the high-accuracy regime. %
\end{itemize}

The paper is organized as follows. After a brief review of related works in Section~\ref{sec:related}, Section~\ref{sec:analysis} presents an analysis covering different aspects that have guided the design of our method. We introduce our approach in Section~\ref{sec:method} and evaluated it in Section~\ref{sec:experiments}. Then we conclude.

\section{Related work}
\label{sec:related}

Consider a set of $N$ elements $\Xset=\{x_1,\dots,x_N\} \subset \Omega$ and a distance $d:\Omega \times \Omega \rightarrow {\mathbb R}$ (or similarity), we tackle the problem of finding the nearest neighbors ${\mathcal N}_{\Xset}(y) \subset \Xset$ of a query $y \in \Omega$, \ie, the elements $\{x\}$ of $\Xset$ minimizing the distance $d(y,x)$ (or maximizing the similarity, respectively). We routinely consider the case $\Omega={\mathbb R}^d$ and $d=\ell_2$, which is of high interest in computer vision applications.

Somehow reminiscent of the research field of compression in the 90\emph{s}, for which we have witnessed a rapid shift from lossless to lossy compression, the recent research effort in this area has focused on \emph{approximate} near- or nearest neighbor search~\cite{IM98,I02,DWJCL08,JDS11,GL11,BL15pami}, in which the guarantee of exactness is traded against high efficiency gains. 

Approximate methods typically improve the efficiency by restricting the distance evaluation to a subset of elements, which are selected based on a locality criterion induced by a space partition. For instance Locality Sensitive Hashing (LSH) schemes~\cite{DIIM04,AI06} exploit the hashing properties resulting from the Johnson-Lindenstrauss lemma. 
Errors occur if a true positive is not part of the selected subset.

Another source of approximation results from compressed representations, which were  pioneered by Weber~\etal~\cite{WSB98} to improve search efficiency~\cite{WSB98}. Subsequently the seminal work~\cite{C02} of Charikar on sketches has popularized compact binary codes as a scalability enabler~\cite{IT03,LCL04}. In these works and subsequent ones employing vector quantization~\cite{JDS11}, errors are induced by the approximation of the distance, which results in swapped elements in the sorted result lists. 
Typically, a vector is reduced by principal component analysis (PCA) dimensionality reduction followed by some form of quantization, such as scalar quantization~\cite{SJ10}, binary quantization~\cite{GP11} and product quantization or its variants~\cite{JDS11,GHKS13}. 
Recent similarity search methods often combine these two approximate and complementary strategies, as initially proposed by J\'egou \etal~\cite{JDS11}. %
The quantization is hierarchical, \ie, a first-level quantizer produces an approximate version of the vector, and an additional code refines this approximation~\cite{JTDA11,BL14}. 

The IVFADC method of~\cite{JDS11} and IMI~\cite{BL15pami} are representative search algorithms employing two quantization levels. All the codes having the same first-level quantization code are stored in a contiguous array, referred to as an inverted list, which is scanned sequentially. 
AnnArbor~\cite{babenko2017annarbor} encodes the vectors w.r.t. a fixed set of nearest vectors. Section~\ref{sec:analysis} shows that this choice is detrimental, and that learning the set of anchor vectors is necessary to reach a good accuracy.

\paragraph{Graph-based approaches.} 
Unlike approaches based on space partitioning, the inspirational NN-descent algorithm~\cite{DCL11} builds a knn-graph to solve the all-neighbors problem: the goal is to find the $k$ nearest neighbors in $\Xset$, w.r.t. $d$, for each $x \in \Xset$. The search procedure proceeds by local updates and is not exhaustive, \ie, the algorithm converges without considering all pairs $(x,x') \in \Xset^2$. 
The authors of NN-decent have also considered it for the approximate nearest neighbor search.

Yuri Malkov~\etal~\cite{malkov2014approximate,malkov2016efficient} introduced the most accomplished version of this algorithm, namely HNSW. This solution selects a series of nested subsets of database vectors, or ``layers''. The first layer contains only a single point, and the base layer is the whole dataset. The sizes of the layers follow a geometric progression, but they are otherwise sampled randomly.
For each of these layers HNSW constructs a neighborhood graph. The search starts from the first layer. A greedy search is performed on the layer until it reaches the nearest neighbor of the query within this layer. That vector is used as an entry point in the next layer as a seed point to perform the search again. At the base layer, which consists of all points, the procedure differs: a bread first search starting at the seed produces the resulting neighbors. 

It is important that the graph associated with each subset %
is \emph{not} the exact knn-graph of this subset: long-range edges must be included. This is akin to simulated annealing or other diversification techniques in optimization: a fraction of the evaluated elements must be far away. In HNSW, this diversification is provided in a natural way, thanks to the long-range links enforced by the upper levels of the structure, which are built on fewer points. However, this is not sufficient, which led Malkov \etal to design a ``shrinking'' operator that reduces a list of neighbors for a vector in a way that does not necessarily keeps the nearest ones. %

\section{Preliminary analysis}
\label{sec:analysis}

This section presents several studies that have guided the design of the approach introduced in Section~\ref{sec:method}. 
All these evaluations are performed on $\Xset= \mathrm{Deep1M} \subset {\mathbb R}^{96}$, \ie, the first million images of the Deep1B dataset~\cite{babenko2016efficient}. 

First, we carry out a comparison between the graph-based traversal of HNSW and the clustering-based hashing scheme employed in IMI. Our goal is to measure how effective a method is at identifying a subset containing neighbors with a minimum number of comparisons. 
Our second analysis considers different estimators of a vector to best approximate it under certain assumptions, including cases where an oracle provides additional information such as the neighbors. 
Finally, we carry out a comparative evaluation of different methods for encoding the descriptors in a compact form, assuming that exhaustive search with approximate representations is possible. This leads us to identify appealing choices for our target operating points.

\subsection{Selectivity: HNSW versus IMI}

We consider two popular approaches for identifying a subset of elements, namely the multi-scale graph traversal of HNSW~\cite{malkov2016efficient} and the space partitioning employed in IMI~\cite{BL15pami}, which relies on a product quantizer~\cite{JDS11}. 
Both methods consists of (i) an identification stage, where the query vector is compared with a relatively small set of vectors (centroids or upper level in HNSW); and (ii) a comparison stage, in which most of the actual distance evaluations are performed. 
For a more direct comparison, 
we compute the exact distances between vectors. We measure the trade-off between accuracy and the number of distance calculations. This is linearly related to selectivity: this metric~\cite{PJA10} measures the fraction of elements that must be looked up.

\begin{figure}[t]
\includegraphics[width=\linewidth]{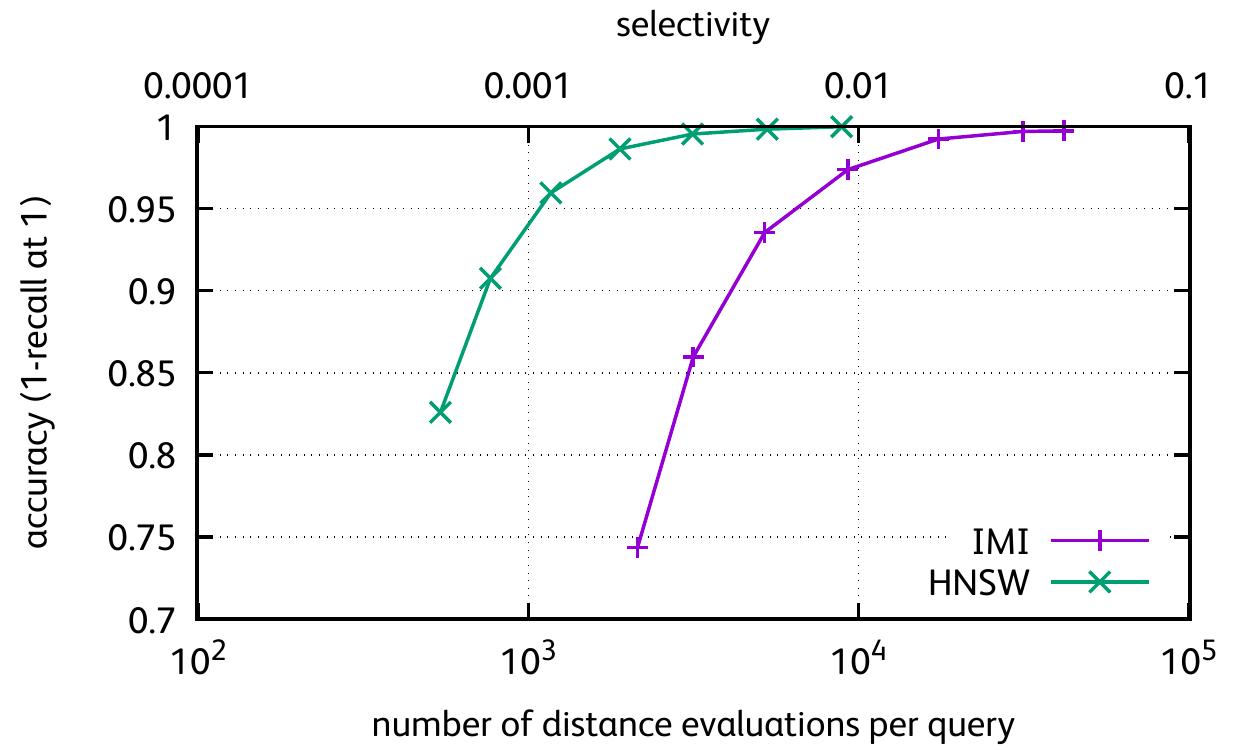}
\caption{\label{fig:distancecomputations}
	IMI \emph{vs} HNSW: accuracy for a given selectivity.
}
\end{figure}

We select  standard settings for this setup: for IMI, we use 2 codebooks of $2^{10}$ centroids, resulting in about 1M inverted lists. For HNSW, we use 64 neighbors on the base layer and 32 neighbors on the other ones. 
During the refinement stage, both methods perform code comparisons starting from most promising candidates, and store the $k$  best search results. The number of comparisons after which the search is stopped is a search-time parameter $T$ in both methods. 
Figure~\ref{fig:distancecomputations} reports the accuracy as a function of the number of distance computations performed for both methods. 

\begin{figure}[t]
\includegraphics[width=\linewidth]{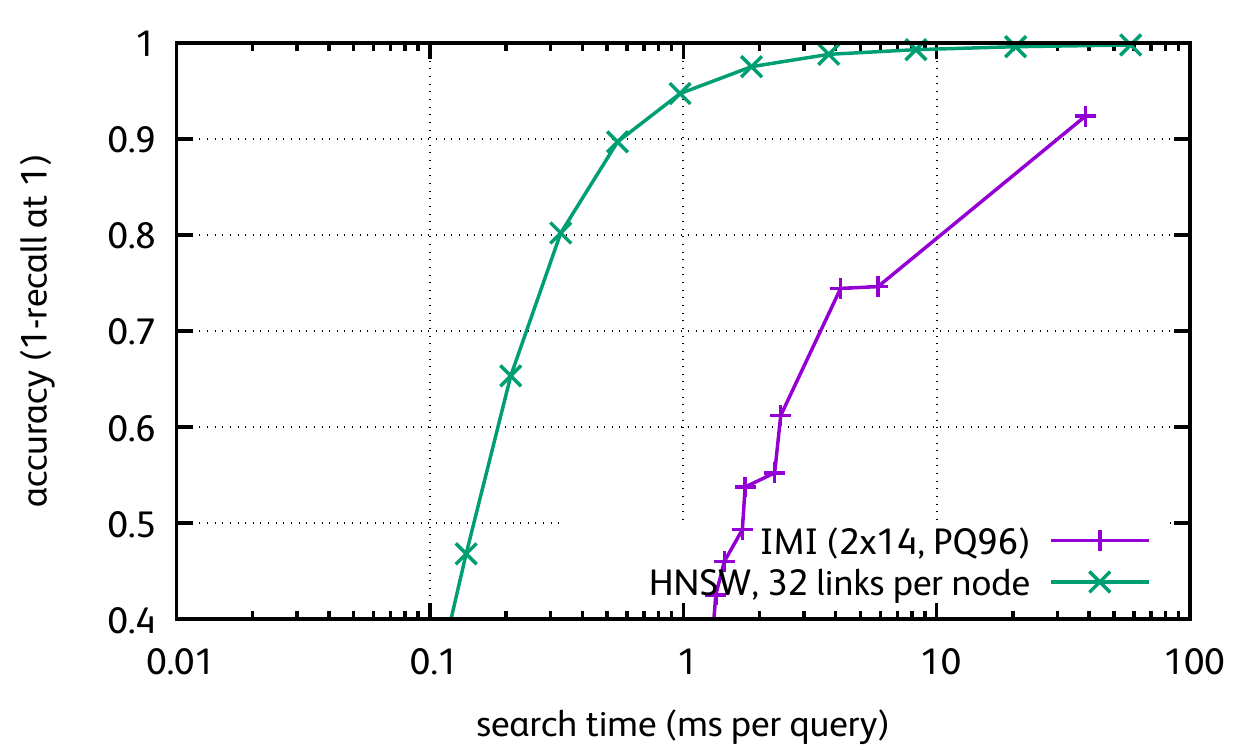}
\caption{\label{fig:midsize}
	IMI \emph{vs} HNSW on mid-sized dataset (Deep100M): trade-off between speed and accuracy. Both methods use 96-byte encodings of descriptors. The HNSW memory size is larger because of the graph connectivity, see text for details. %
}
\end{figure}

The plot shows that HNSW is 5 to 8 times more selective than IMI for a desired level of accuracy. 
This better selectivity does not directly translate to the same speed-up because HNSW requires many random probes from main memory, as opposed to contiguous inverted lists. Yet this shows that \emph{HNSW will become invariably better than IMI for larger vector representations}, when the penalty of random accesses does not dominate the search time anymore. Figure~\ref{fig:midsize} confirms that HNSW with a scalar quantizer is faster and more accurate than an IMI employing very fine quantizers at both levels. However, this requires 224~bytes per vector, which translates to 50~GB in RAM when including all overheads of the data structure.

\subsection{Centroids, neighbor or regression?}
\label{sec:approximators}

Hereafter we investigate several ways of getting a coarse approximation\footnote{Some of these estimations depend on additional information, for example when we assume that all other vertices of a given database vector $x$ are available. We report them as topline results for the sake of our study.}  of a vector $x \in \Xset$:
\begin{description}
\item [Centroid.] We learn by k-means a coarse codebook ${\mathcal C}$ comprising 16k elements. It is learned either directly on $\Xset$ or using a distinct training set of 1 million vectors. We approximate $x$ by its nearest neighbor $q(x) \in {\mathcal C}$. 

\item [Nearest neighbor.] We assume that we know the nearest neighbor $n_1(x)$ of $x$ and can use it as an approximation. This choice %
shows the upper bound of what we can achieve by selecting a single vector in $\Xset$. 

\item [Weighted average.] Here we assume that we have access to the $k=8$ nearest neighbors of $x$ ordered by decreasing distances, stored in matrix form as $\N(x)=[n_1,\dots,n_k]$. We estimate $x$ as the weighted average 
\begin{equation}
\bar{x} = \betastar^\top \N(x), 
\end{equation}
where $\betastar$ is a fixed weight vector constant shared by all elements in $\Xset$.  
The close-form computation of $\betastar$ is detailled in Section~\ref{sec:method}. 

\item [Regression.] Again we use $\N(x)$ to estimate $x$, but we additionally assume that we perfectly know  the optimal regression coefficients $\beta(x)$ minimizing the reconstruction error of $x$. In other words we compute 
\begin{equation}
\hat{x} = \beta(x)^\top \N(x), 
\label{equ:xhat}
\end{equation}
where $\beta(x)$ is obtained as the least-square minimizer of the over-determined system $\| x - \beta(x)^\top \N(x) \|^2$. 
\end{description}

\begin{figure}[t]
\includegraphics[width=\linewidth]{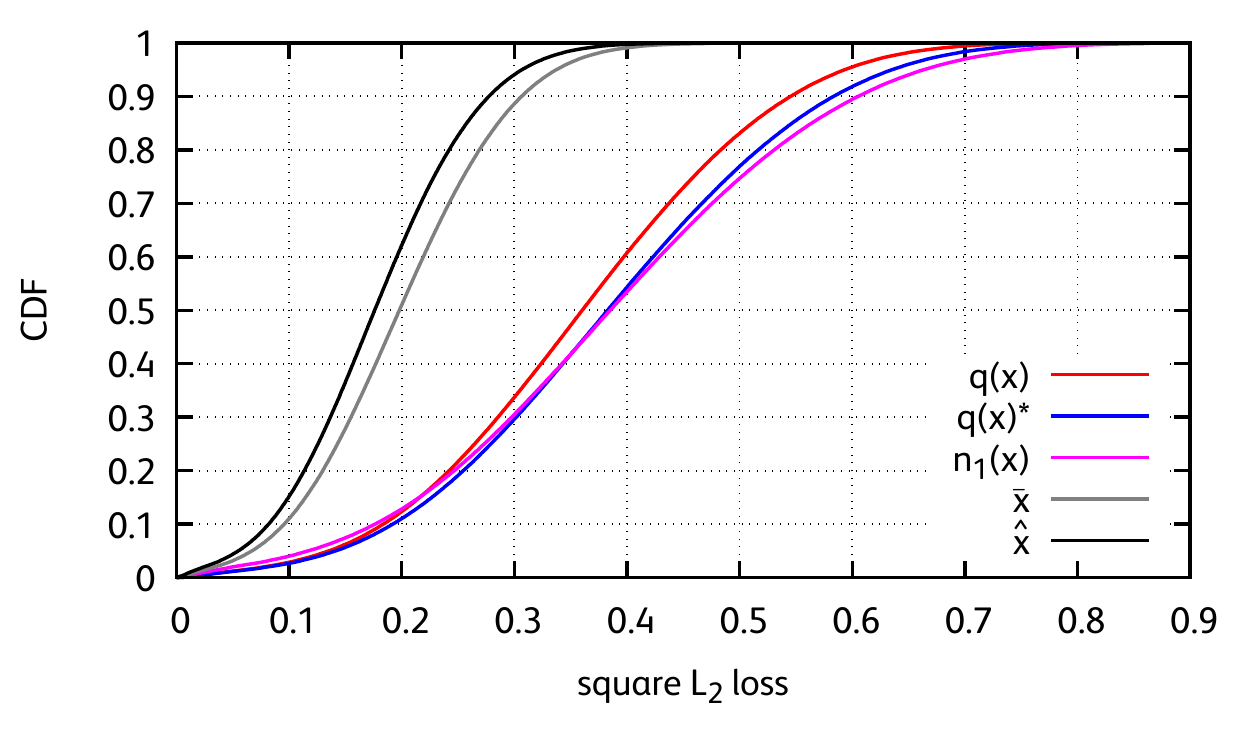}
\caption{Cumulative probability mass function of the square loss for different estimators of $x$. $q(x)$: the closest centroid from a codebook learned on $\Xset$ or ($q(x)^*$) a distinct set;  $n_1(x)$: nearest neighbor in $\Xset$; $\bar{x}$: weighted average of 8 neighbors; $\hat{x}$: the best estimator from its 8 nearest neighbors. 
\label{fig:L2sqr}}
\end{figure}

Figure~\ref{fig:L2sqr} shows the distribution of the error (square Euclidean distance) for the different estimators. 
We draw several observations. First, choosing the centroid $q(x)$ in a codebook of 16k vectors is comparatively more effective than taking the nearest neighbor $n_1(x)$ amongst the $64\times$ larger set $\Xset$ of 1 million vectors.  
Therefore using vectors of the HNSW upper level graph as reference points to compute a residual vector is not an interesting strategy. 

Second, if the connectivity is granted for free or required by design like in HNSW, the performance achieved by $\bar{x}$ suggests that we can improve the estimation of $x$ from its neighbors with no extra storage,  if we have a reasonable approximation of $N(x)$. %

Third, under the same hypotheses and assuming additionally that we have the parameter $\beta(x)$ for all $x$, a better estimator can be obtained with Eqn.~\ref{equ:xhat}. This observation is the key to the re-ranking strategy introduced in Section~\ref{sec:method}.

\subsection{Coding method: first approximation}

We evaluate which vector compression is most accurate \emph{per se} to perform an initial search given a memory budget. Many studies of this kind focus on very compact codes, like 8 or 16 bytes per vector. We are interested in higher-accuracy operating points. Additionally, the results are often reported for systems parametrized by several factors (short-list, number of probes, etc), which makes it difficult to separate the false negatives induced by the coding from those resulting from the search procedure. 

To circumvent this comparison issue, we compress and decompress the database vectors, and perform an exhaustive search.  
All experiments in Table~\ref{tab:compression} are performed on Deep1M (the 1M first images of the Deep1B dataset). The codebooks are trained on the provided distinct training set. We consider in particular product quantization (PQ~\cite{JDS11}) and optimized product quantizer (OPQ~\cite{GHKS13}). We adopt a notation of the form PQ16x8 or OPQ14x2, in which the values respectively indicate the number of codebooks and the number of bits per subquantizer. 

Additionally, we consider a combination of quantizers exploiting residual vectors~\cite{JDS11,JTDA11} to achieve higher-accuracy performance. In this case, we consider a 2-level residual codec, in which the first level is either a vector quantizer with 65536 centroids (denoted PQ1x16 in our notation) or a product quantizer (PQ2x12 or PQ2x14). What remains of the memory budget is used to store a OPQ code for the refinement codec, which encodes the residual vector. Note that IVFADC-based methods and variants like IMI~\cite{BL12} exploit 2-level codecs. Only the data structure differs. 

Our results show that 2-level codecs are more accurate than 1-level codecs. They are also more computationally expensive to decode. 
For operating points of 32 bytes, we observe that just reducing the vectors by PCA or encoding them with a scalar quantizer is sub-optimal in terms of accuracy. Using OPQ gives a much higher accuracy. Thanks to the search based on table lookups, it is also faster than a scalar quantizer in typical settings. 
For comparison with the AnnArbor method, we also report a few results on 16 bytes per vector. The same conclusions hold: a simple 2-level codec with 65536 centroids (\eg, PQ1x16+OPQ14x8) gets the same codec performance as AnnArbor. 

\begin{table}
\centering
{\small
\begin{tabular}{lrll}
codec & size\ \ \  & \multicolumn{2}{c}{accuracy} \\
      & (bytes)  &  \hspace{-4pt}recall@1 & \hspace{-4pt}recall@10\\
\hline
none               & 384   & 1.000 & 1.000    \\
scalar quantizer   & 96    & 0.978 & 1.000 \\
\hline
PQ16x8             & 16    & 0.335 & 0.818 \\
PQ8x16             & 16    & 0.394 & 0.881 \\
PQ2x8+OPQ14x8      & 16    & 0.375 & 0.867 \\
PQ1x16+OPQ14x8 
                   & 16    & 0.422 & 0.899 \\
PQ2x16+OPQ12x8     & 16    & 0.421 & 0.904 \\
PQ2x12+OPQ13x8     & 16    & 0.382 & 0.870 \\
AnnArbor~\cite{babenko2017annarbor}   
                   & (*)\ \  16 &  0.421  &   \\
\hline	
OPQ32x8             	& 32   & 0.604 & 0.982\\
PQ1x16+OPQ30x8 	& 32   & 0.731 & 0.997 \\
PQ2x16+OPQ28x8	& 32   & 0.713 & 0.996 \\
PQ2x14+OPQ28x8	& 32   & 0.693 & 0.995 \\
PCA8			& 32   & 0.017 & 0.074 \\
\hline
\end{tabular}}
\smallskip
\caption{\label{tab:compression}
    HNSW: Exhaustive search in 1M vectors in 96D, with different coding methods. We report the percentage of queries for which the nearest neighbor is among the top1 (resp. top10) results. \newline (*) AnnArbor depends on a parenthood link (4 bytes).
}
\end{table}

\section{Our approach: \OURMETHOD}
\label{sec:method}

This section describes our approach, namely \OURMETHOD (link and code). It offers a state-of-the-art compromise between approaches considering very short codes (8--32 bytes) and those not considering the memory constraint, like FLANN and HNSW. 
After presenting an overview of our indexing structure and search procedure, we show how to improve the reconstruction of an indexed vector from its approximate neighbors with no additional memory. Then we introduce our novel refinement procedure with quantized regression coefficients, and details the optimization procedure used to learn the regression codebook. We finally conduct an analysis to discuss the trade-off between connectivity and coding, when fixing the memory footprint per vector.

\subsection{Overview of the index and search}

\paragraph{Vector approximation.} All indexed vectors are first compressed with a coding method independent of the structure. It is a quantizer, which formally maps any vector $x \in {\mathbb R}^d \mapsto q(x) \in {\mathcal C}$, where ${\mathcal C}$  is  a finite subset of ${\mathbb R}^d$, meaning that $q(x)$ is stored as a code. 

Following our findings of Section~\ref{sec:analysis}, we adopt two-level encodings for all experiments. For the first level, we choose a product quantizer of size 2x12 or 2x14 bits (PQ2x12 and PQ2x14), which are cheaper to compute. For the second level, we use OPQ with codes of arbitrary length.

\paragraph{Graph-based structure.} We adopt the HSNW indexing structure, except that we modify it so that it works with our coded vectors. More precisely, all vectors are stored in coded format, but the \texttt{add} and \texttt{query} operations are performed using asymmetric distance computations~\cite{JDS11}: the query or vector to insert is not quantized, only the elements already indexed are. We fix the degree of the graphs at the upper levels to $k=32$, and the size ratio between two graph levels at 30, \ie, there are 30$\times$ fewer elements in the graph level 1 than in the graph level 0.

\paragraph{Refinement strategy.} We routinely adopt a two-stage search strategy~\cite{JTDA11}. During the first stage, we solely rely on the first approximation induced by $q(\cdot)$ to select a short-list of potential neighbor candidates. The indexed vectors are reconstructed on-the-fly from their compact codes. 
The second stage requires more computation per vector is and applied only on this short-list to re-rank the candidates. 
We propose two variants for this refinement procedure:
\begin{itemize}
\item Our \textbf{0-byte} refinement does not require any additional storage per vector. It is performed by re-estimating the candidate element from its connected neighbors encoded with $q(x)$. Section~\ref{sec:0byte} details this method. 
\item We refine the vector approximation by using a set of quantized regression coefficients stored for each vector. These coefficients are learned and selected for each indexed vector offline, at building time, see Section~\ref{sec:regressNN}. 
\end{itemize}

\subsection{0-byte refinement} 
\label{sec:0byte}

Each indexed vector $x$ is connected in the graph to a set of $k$ other vectors, $g_1(x),\dots,g_k(x)$, ordered by increasing distance to $x$. This set can include some of the nearest neighbors of $x$, but not necessarily. From their codes, we reconstruct $x$ as $q(x)$ and each $g_i$ as $q(g_i(x))$. 
We define the matrix
$\G(x) = \left[q(x), q(g_1(x)), \dots, q(g_k(x))\right]$
stacking the reconstructed vectors. Our objective is to use this matrix to design a better estimator of $x$ than $q(x)$, \ie, to minimize the expected square reconstruction loss. For this purpose, we minimize the empirical loss
\begin{equation}
L(\beta) = \sum_{x \in \Xset}  \| x - \beta^\top \G(x) \|^2.
\label{equ:betastar}
\end{equation}
over $\Xset$. Note that, considering the small set of $k+1$ parameters, using a subset of $\Xset$ does not make any difference in practice. We introduce the vertically concatenated vector and matrix  %
\vspace{-10pt}
\begin{equation}
\X=
\begin{bmatrix}
x_1 \\
\vdots \\
x_N \\
\end{bmatrix}
\quad \textrm{and} \quad
\Y = 
\begin{bmatrix}
\G(x_1) \\
\vdots \\
\G(x_N) \\
\end{bmatrix}
\end{equation}
and point out that $L(\beta) = \| X - \beta^\top \Y \|^2$.
This is a regular least-square problem with a closed-form solution $\betastar=\Y^*X$, where $\Y^*$ is the Moore-Penrose pseudo-inverse of $\Y$. We compute the minimizer $\betastar$ with a standard regressor. 
This regression weights are shared by all index elements, and therefore do not involve any per-vector code. 
A indexed vector is refined from the compact codes associated with $x$ and its connected vectors as 
\begin{equation}
\bar{x} = \betastar^\top \G(x). 
\end{equation}

In expectation and by design, $\bar{x}$ is a better approximation of $x$ than $q(x)$, \ie, it reduces the quantization error. 
It is interesting to look at the weight coefficient in $\betastar$ corresponding to the vector $q(x)$ in the final approximation. It can be as small as $0.5$ if the quantizer is very coarse: in this situation the quantization error is large and we significantly reduce it by exploiting the neighbors. In contrast, if the quantization error is limited, the weight is typically $0.9$. 
\medskip

\subsection{Regression codebook}
\label{sec:regressNN}

The proposed 0-byte refinement step is granted for free, given that we have a graph connecting each indexed element with nearby points. As discussed in Section~\ref{sec:approximators}, a vector $x$ would be better approximated from its neighbors if we knew the optimal regression coefficients. This requires to store them on a per-vector basis, which would  increase the memory footprint per vector by $4\times k$ bytes with floating-point values. 
In order to limit the additional memory overhead, we now describe a method to learn a codebook $\Bset=\{\beta_1,\dots,\beta_B\}$ of regression weight vectors. Our objective is to minimize the empirical loss
\begin{equation}
L'(\Bset) = \sum_{x \in \Xset} \min_{\beta \in \Bset} \|x - \beta^\top \G(x)\|^2. 
\label{equ:lossB}
\end{equation}

Performing a k-means directly on regression weight vectors would optimize the $\ell_2$-reconstruction of the regression vector $\beta(x)$, but not of the loss in Eqn.~\ref{equ:lossB}. We use k-means only to initialize the regression codebook. Then we use an EM-like algorithm alternating over the two following steps. 
\begin{enumerate}
\item \textbf{Assignment.} Each vector $x$ is assigned to the codebook element minimizing its reconstruction error: 
\begin{equation}
\beta(x) = \arg \min_{\beta \in \Bset} \|x - \beta^\top \G(x)\|^2.
\end{equation}

\item \textbf{Update.} For each cluster, that we conveniently identify by $\beta_i$, we find the optimal regression weights 
\begin{equation}
\beta_i^{\star} = \arg \min_{\beta} \sum_{x \in \Xset: \beta(x)=\beta_i}  \| x - \beta^\top \G(x) \|^2 
\label{equ:betacluster}
\end{equation}
and update $\beta_i \leftarrow \beta_i^{\star}$ accordingly. 
\end{enumerate}
For a given cluster, Eqn.~\ref{equ:betacluster} is the same as the one of Eqn.~\ref{equ:betastar}, except that the solution is computed only over the subset of vectors assigned to $\beta_i$. 
It is  closed-form as discussed earlier. 

In practice, as $B$ is relatively small ($B=256$), we only need a subset of $\Xset$ to learn a representative codebook~$\Bset$. %
This refinement stage requires 1 byte per indexed vector to store the selected weight vector from the codebook~$\Bset$. 

\vspace{-1em}
\paragraph{Product codebook.} As  shown later in the experimental section, the performance improvement brought by this regression codebook is worth the extra memory per vector. However, the performance rapidly saturates as we increase the codebook size~$B$. This is expected because the estimator $\beta(x)^\top \G(x)$ only spans a $(k+1)$-dimensional subspace of ${\mathbb R}^d$, $k \ll d$. 
Therefore the projection of $x$ lying in the null space $\mathrm{ker}(\G)$ cannot be recovered. 

We circumvent this problem by adopting a strategy inspired by product quantization~\cite{JDS11}. We evenly split each vector as $x=[x^1;\dots,x^M]$, where each $x^j \in \mathbb{R}^{d/M}$, and learn a product regression codebook $\Bset^1\times ... \times \Bset^M$, \ie, one codebook per subspace. In this case, extending the superscript notation to $\beta$ and $\G$, the vector is estimated as 
\begin{equation}
\hat{x} = [\beta^1(x^1) \G^1(x), \dots, \beta^M(x^M) \G^M(x)], 
\end{equation}
where $\forall j,\ \beta^j(x^j) \in \Bset^j$. 
The set of possible estimators spans a subspace having to up $M\times k$ dimensions. 
This refinement method requires $M$ bytes per vector.

\subsection{\OURMETHOD: memory/accuracy trade-offs}

As discussed earlier, HNSW method is both the fastest and most accurate indexing method at the time being, but its scalability is restricted by its high memory usage. For this reason, it has never been demonstrated to work at a billion-scale. 
In this subsection, we analyze our algorithm \OURMETHOD when imposing \emph{a fixed memory budget per vector}. Three factors contribute to the marginal memory footprint:
\begin{itemize}
\item the code used for the initial vector approximation, for instance OPQ32 (32 bytes);
\item the number $k$ of graph links per vector (4 bytes per link);
\item \xspace[\emph{optionally}] the $M$ bytes used by our refinement method from neighbors with a product regression codebook. 
\end{itemize}

\paragraph{\OURMETHOD Notation.} To identify unambiguously the parameter setting, we adopt a notation of the form L6\&OPQ40. L6 indicates that we use 6 links per vector in the graph and OPQ40 indicates that we use first encode the vector with OPQ, allocating  $40$ bytes per vector. If, optionally, we use a regression codebook, we refer to as by the notation M=4 in tables and figures. The  case of 0-coding is denoted by M=0. 

\begin{figure}
\includegraphics[width=\linewidth]{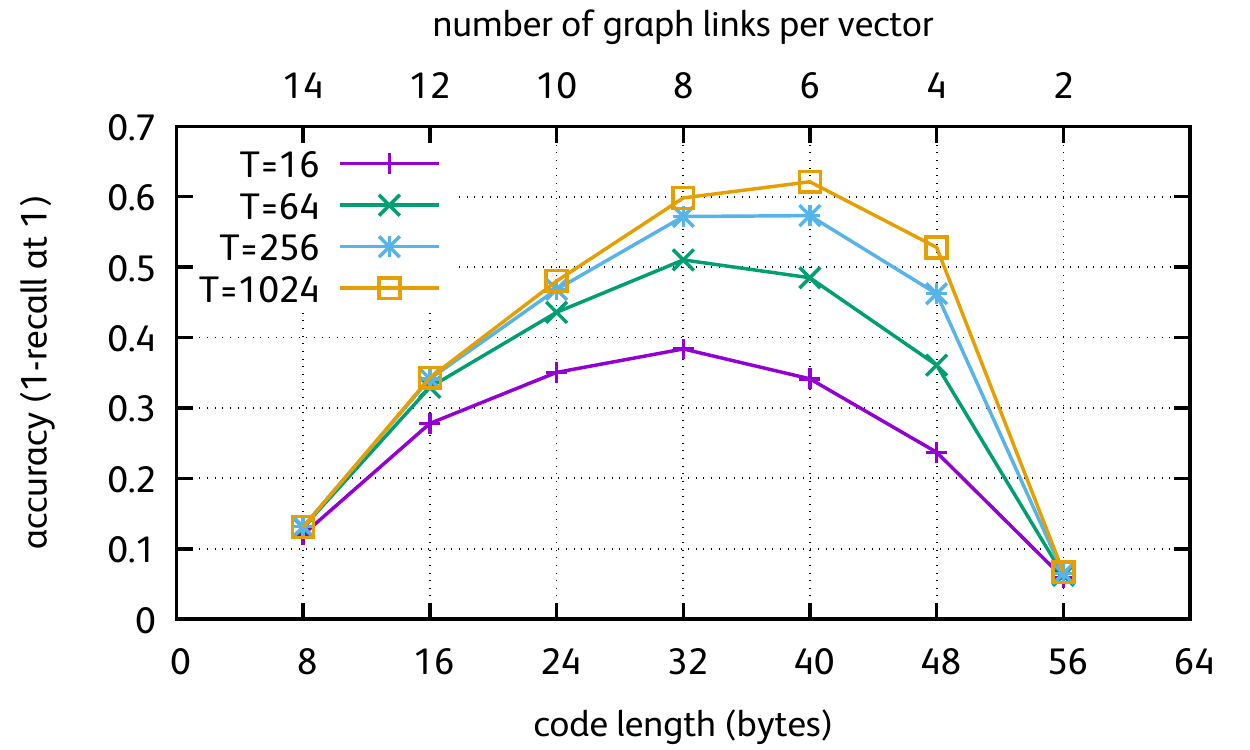}
\caption{\label{fig:basictradeoff}
	Deep1M: Performance obtained depending on whether we allocate a fixed memory budget of 64 bytes to codes (OPQ codes of varying size) or links. Recall that $T$ is the parameter capping the number of distance evaluations. 
}
\end{figure}

\vspace{-1em}
\paragraph{Coding \emph{vs} Linking.} We first consider the compromise between the number of links and the number of bytes allocated to the compression codec. 
Figure~\ref{fig:basictradeoff} is a simple experiment where we start from the full HNSW representation and reduce either the number of links or the number of dimensions stored for the vectors. We consider all setups reaching the same budget of 64 bytes, and report results for several choices of the parameter $T$, which controls the total number of comparisons. 

We observe that there is a clear trade-off enforced by the memory constraint. The search is ineffective with too few links, as the algorithm can not reache all points. At the opposite side, the accuracy is also impacted by a too strong approximation of the vector, when the memory budget allocated to compression is insufficient. Interestingly, increasing $T$ shifts the optimal trade-off towards allocating more bytes to the code. This means that the neighbors can be reached but require more hops in the graphs.  

\vspace{-1em}
\paragraph{Coding vectors \emph{vs} regression coefficients.}
We now fix the number of links to 6 and evaluate the refinement strategy under a fixed total memory constraint. In this case we have a trade-off between the number of bytes allocated to the compression codec and to the refinement procedure. 

The first observation drawn from Table~\ref{tab:codetradeoffs} is that the two refinement methods proposed in this section both significantly reduce the total square loss. This behavior is expected for the 0-coding because it is exactly what the method optimizes. However, this better reconstruction performance does not translate to a better recall in this setup. We have investigated the reason of this observation, and discovered that the 0-coding approach gives a clear gain when regressing with the exact neighbors, but those provided by the graph structure have more long-range links. 

In contrast, our second refinement strategy is very effective. Coding the regression coefficients with our codebook significantly improves both the reconstruction loss and the recall: the refinement coding based on the graph is more effective than the first-level coding, which is agnostic of the local distribution of the vectors. 

\begin{table}
\hspace*{-5pt}
{\footnotesize
\begin{tabular}{@{}l@{\mysp}|@{\mysp}c@{\mysp}c@{\mysp}|@{\mysp}c@{\mysp}c@{\mysp}|@{\mysp}c@{\mysp}c@{}}
                     & \multicolumn{2}{@{\mysp}c@{\mysp}|@{\mysp}}{vector quantization}  & \multicolumn{4}{@{\mysp}c@{\mysp}}{R@1} \\
\ \ \ codec          & \multicolumn{2}{@{\mysp}c@{\mysp}|@{\mysp}}{error ($\times 10^3$)} & \multicolumn{2}{@{\mysp}c@{\mysp}|@{\mysp}}{exhaustive}  & T=1024 & T=16384\\
\hline
\hspace{35pt} Deep:  & 100M & 1B & 100M & 1B & 100M & 1B \\
\hline
L6\&OPQ40           	& 24.3  & 24.3	& 0.608	& 0.601	& 0.427	& 0.434\\
L6\&OPQ40 M=0    	& 22.7  & 22.5	& 0.611	& 0.600	& 0.429	& 0.435\\
L6\&OPQ36 M=4     	& 21.9  & 21.5 	& 0.608	& 0.607	& 0.428	& 0.434\\
L6\&OPQ32 M=8    	& 20.0  & 19.8 	& 0.625	& 0.612	& 0.438	& 0.438\\
\hline
\end{tabular}}
\smallskip
\caption{\label{tab:codetradeoffs}
Under a constraint of 64 bytes and using $k=6$ links per indexed vector, we consider different trade-offs for allocating bytes to between codes for reconstruction and neighbors. %
}
\vspace{-1em}
\end{table}

\section{Experiments}
\label{sec:experiments}

The experiments generally evaluate the search time \emph{vs} accuracy tradeoff, considering also the size of the vector representation. The accuracy is measured as the fraction of cases where the actual nearest neighbor of the query is returned at rank 1 or before some other rank (recall @ rank). The search time is given in milliseconds per query on a 2.5~GHz server with 1~thread. Parallelizing searches with multiple threads is trivial but timings are less reproducible.

\subsection{Baselines \& implementation}

We chose IMI as a baseline method because most recent works on large-scale indexing build upon it~\cite{KA14,BL14a,babenko2016efficient,douze2016polysemous} and top results for billion-scale search are reported by methods relying on it. We use the competitive implementation of Faiss~\cite{johnson2017billion} (in CPU mode) as the IMI baseline. We use the automatic hyperparameter tuning to optimize IMI's operating points. The parameters are the number of visited codes (T), the multiprobe number and the Hamming threshold used to compare polysemous codes~\cite{douze2016polysemous}. 

Our implementation of HNSW follows the original NMSLIB version~\cite{BN13}. The most noticeable differences are that (i) vectors are added by batches because the full set of vectors does not fit in RAM, and (ii) the HNSW structure is built layer by layer, which from our observation improve the quality of the graph. Indexing 1~billion vectors takes about 26~hours with \OURMETHOD: we can add more than 10,000 vectors per second to the index. We refine at most 10 vectors. 

For the encodings, we systematically perform a rotation estimated with Optimized Product Quantization (OPQ) to facilitate the encoding in the second level product quantizer.

\subsection{Large-scale evaluation}

\begin{figure}
\includegraphics[width=0.9\linewidth]{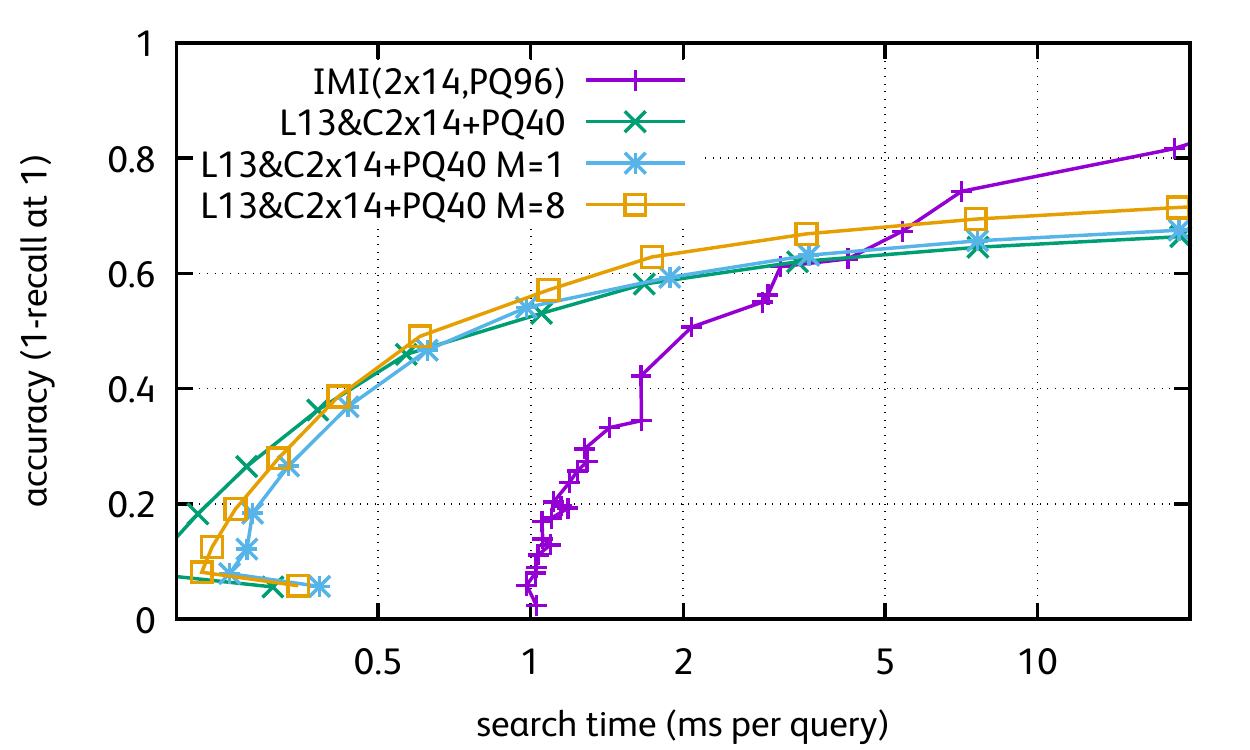}
\includegraphics[width=0.9\linewidth]{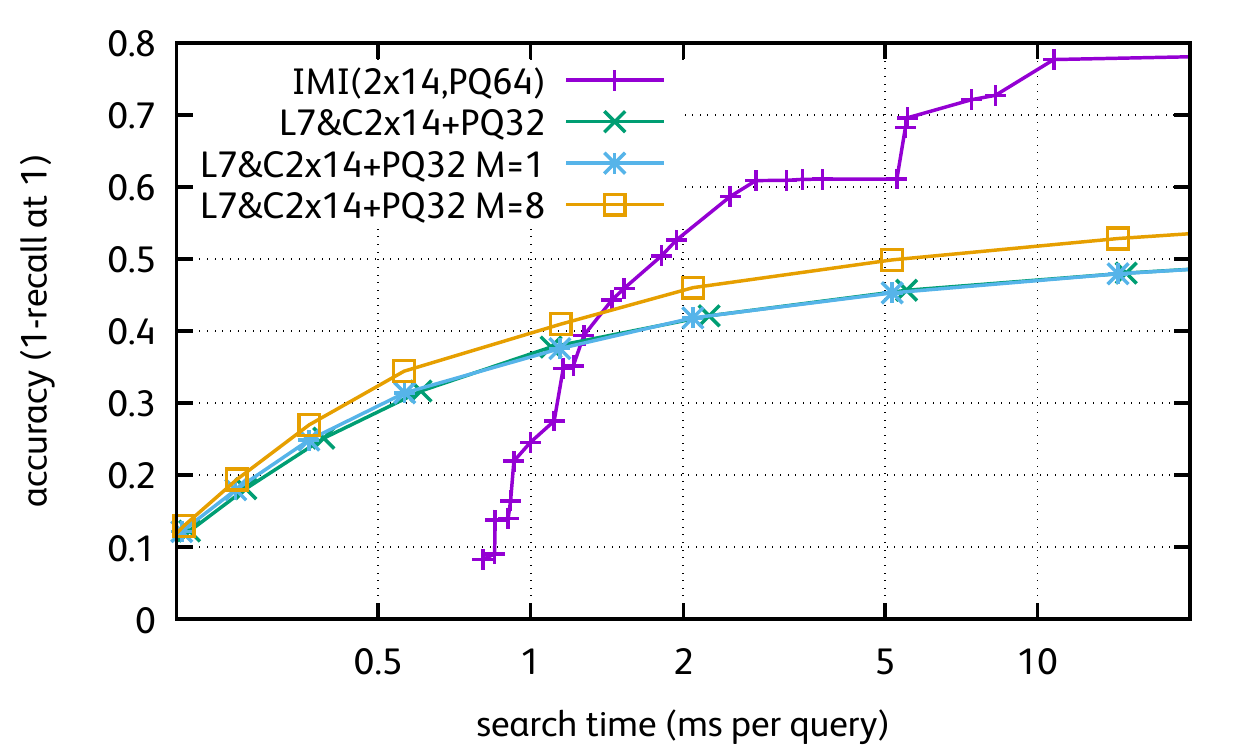} 
\caption{ 
	Speed \emph{vs} accuracy on Deep1B (top) and BIGANN (bottom). Timings are measured per query on 1 core. 
\label{fig:eval1B}}
\vspace{-7pt}
\end{figure}

We evaluate on two large datasets widely adopted by the computer vision community. BIGANN~\cite{JTDA11} is a dataset of 1B SIFT vectors in 256 dimensions, and Deep1B is a dataset of image descriptors extracted by a CNN. Both datasets come with a set of 10,000 query vectors, for which the ground-truth nearest neighbors are provided as well as a set of unrelated training vectors that we use to learn the codebooks for the quantizers. IMI codebooks are trained using 2 million vectors, and the regression codebooks of \OURMETHOD are trained using 250k vectors and 10 iterations.

Figure~\ref{fig:eval1B} compares the operating points in terms of search time \emph{vs} accuracy on Deep1B for encodings that use 96~bytes per vector. For most operating points, our L\&S method is much faster, for example 2.5$\times$ faster to attain a recall@1 of 50\% on Deep1B. The improvement due to the refinement step, \ie the regression from neighborhood, is also significant. It consumes a few more bytes per vector (up to 8). 

For computationally expensive operating points, IMI is better for recall@1 because the $4k$=52 bytes spent for links could be used to represent the vertices more accurately.

\subsection{Comparison with the state of the art}

We compare L\&C with other results reported in the literature, see Table~\ref{tab:SOTA}. Our method uses significantly more memory than others that are primarily focusing on optimizing the compromise between memory and accuracy. However, unlike HNSW, it easily scales to 1 billion vectors on one server. L\&C is competitive when the time budget is small and one is interested by higher accuracy. The competing methods are either much slower, or significantly less accurate. On Deep1B, only the polysemous codes attain an efficienty similar to ours, obtained with a shorter memory footprint. However it only attains recall@1=45.6\%, against 66.7\% for \OURMETHOD. 
Considering the recall@1, we outperform the state of the art on BIGANN by a large margin with respect to the accuracy/speed trade-off. 

Note, increasing the coding size with other methods would increase accuracy, but would also invariably increase the search time. 
Considering that, in a general application, our memory footprint remains equivalent or smaller than other meta-data associated with images, our approach offers an appealing and practical solution in most applications.

\begin{table}
\scalebox{0.8}{
\begin{tabular}{@{\hspace{0pt}}l@{\hspace{7pt}}rrrrr@{\hspace{0pt}}}
\hline
                    & R@1   &  R@10 & R@100 & time (ms)  & bytes \\
\hline 
\multicolumn{6}{c}{BIGANN} \\
Multi-LOPQ~\cite{KA14} 
                    & 0.430 & \textbf{0.761} & 0.782 &  \texttt{8\ \ \ }          & 16 \\                     
OMulti-D-OADC-L~\cite{BL15pami} &  0.421& 0.755 & 0.782 & \texttt{7\ \ \ }        & 16 \\

FBPQ \cite{BL14a}   & 0.179 & 0.523 & 0.757 &  \texttt{1.9\ }    & 16  \\
                    & 0.186 & 0.556 & \textbf{0.894} &  \texttt{9.7\ }       & 16 \\
Polysemous~\cite{douze2016polysemous}
                    & 0.330 &       & 0.856 & \texttt{2.77}       & 16 \\
L7\&C32 M=8           & \textbf{0.461} & 0.608 & 0.613 & \texttt{2.10}       & 72 \\ %
\hline
\multicolumn{6}{c}{Deep1B} \\
GNO-IMI~\cite{babenko2016efficient} 
                    & 0.450  & 0.8   &       & \texttt{20\ \ \ }         & 16 \\ 
Polysemous~\cite{douze2016polysemous}
	                & 0.456 &       &       & \texttt{3.66}       & 20 \\ 
L13\&C40 M=8        & \textbf{0.668} & \textbf{0.826} & \textbf{0.830} & \texttt{3.50}       & 108 \\ %
\hline 
\end{tabular}
} %
\smallskip
\caption{\label{tab:SOTA}State of the art on two billion-sized datasets.
}
\vspace{-5pt}
\end{table}

\section{Conclusion}
\label{sec:conclusion}

We have introduced a method for precise approximate nearest neighbor search in billion-sized datasets. It targets the high-accuracy regime, which is important for a vast number of applications. 
Our approach makes the bridge between the successful compressed-domain and graph-based approaches. The graph-based candidate generation offers a higher selectivity than the traditional structures based on inverted lists. The compressed-domain search allows us to scale to billion of vectors on a vanilla server. 
As a key novelty, we show that the graph structure can be used to improve the distance estimation for a moderate or even null memory budget. 
As a result, we report state-of-the-art results on two public billion-sized benchmarks in the high-accuracy regime. 

Our approach is open-sourced in the Faiss library, see \texttt{https://github.com/facebookresearch/faiss/}
\texttt{tree/master/benchs/}\verb|link_and_code|.

{\small
\bibliographystyle{ieee}
\bibliography{egbib}
}

\end{document}